\documentclass{ecai}
\usepackage{times}
\usepackage{graphicx}
\usepackage{latexsym}

\begin{document}

\title{Verbal Programming of Robot Behavior}

\author{Jonathan Connell\institute{IBM T.J. Watson Research, USA, email: jconnell@us.ibm.com} }

\maketitle
\bibliographystyle{ecai}

\begin{abstract}
Home robots may come with many sophisticated built-in abilities, however there will always be a degree of customization needed for each user and environment. Ideally this should be accomplished through one-shot learning, as collecting the large number of examples needed for statistical inference is tedious. A particularly appealing approach is to simply explain to the robot, via speech, what it should be doing. In this paper we describe the ALIA cognitive architecture that is able to effectively incorporate user-supplied advice and prohibitions in this manner. The functioning of the implemented system on a small robot is illustrated by an associated video \cite{video}.
\end{abstract}

\section{INTRODUCTION}
%The traditional page limit for ECAI long papers is {\bf 7 (six)} pages
%in the required format. The traditional page limit for short
%submissions is {\bf 2} pages.
%
%However, these page limits may change from one ECAI to
%another. Consult the most recent Call For Papers (CFP) for the most
%up-to-date page limits.

A typical home robot of the future might have built-in navigation, object recognition, task planning, and dexterous manipulation. Yet, despite these sophisticated capabilities, there are still things it cannot know when it first arrives. For instance, what a particular room in the house is called, even if it can identify the general type. Is this Freddy's bedroom or Janey's? Is the room with the couch the ``living room", ``den", ``parlor", or ``TV room"? And what objects are in the house, such as ``Pookie", Janey's favorite stuffed animal?  

Beyond naming, there are also the tasks that the owner wants performed. For instance, depending on the user, ``tidy up" might mean:

\begin{itemize}
\item shove everything under the bed
\item throw out all the food wrappers then stuff everything under the bed
\item return the dirty dishes to the kitchen and put the clothes in the hamper
\item carefully put all the toys back into their standard places
\end{itemize}

\noindent How does one impart this knowledge to the robot so it can be maximally useful?

The solution pursued in the ALIA architecture is to just explain these things in English, something that can hopefully be done by a non-technical user. Arguably there are facets of tasks that are difficult to put into words, such as particular locations, paths, or forces to be applied. These might be better handled through pointing \cite{Eli}, demonstration/imitation \cite{PbD,imitation}, and puppeting \cite{compliant,Pook}, respectively. Yet, language is often adequate to describe the bulk of a task, and it is still valuable as glue to hold the non-verbal elements together. Such learning-by-being-told may appear trivial, but it is tricky to implement in practice. In particular, the primitives and control structures in the reasoning system must be tailored to reflect the constructs used by language in order to allow smooth translation. In a rough sense, the ALIA system strives to be an Advice Taker as envisioned in \cite{advice}.

To reason, ALIA builds on a kernel of grounding functions. Some of these link to action-like procedures, such as setting the speed of the wheels or configuring the manipulator. Others invoke sensory analysis routines, like finding all red objects or determining if there is a collision-free path to some destination. The kernel can be thought of as the fresh-from-the-factory capabilities of the base platform. ALIA then allows programs of more complexity to be created on top of these libraries using natural language utterances to string them together.

\begin{figure*}[t]
\centering
\includegraphics[width=0.95\textwidth]{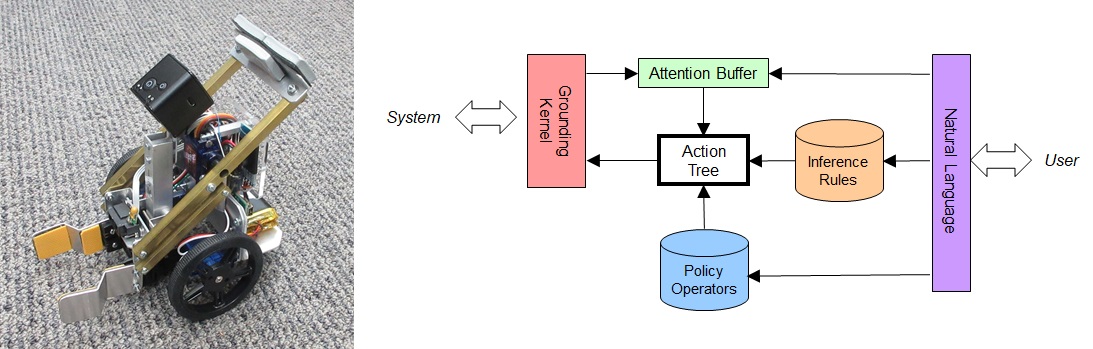} % Reduce the figure size so that it is slightly narrower than the column. Don't use precise values for figure width.This setup will avoid overfull boxes. 
\caption{The ALIA reasoning system uses a set of rules to interpret situations, and a set of operators to generate behaviors. The output of the controller relies on a set of grounding functions intrinsic to the deployment platform, in this case a small forklift robot.}
\label{fig:block}
\end{figure*}

\section{RELATED WORK}

\noindent There is a long history of programming by demonstration in robotics \cite{LFD}. Some systems attempt to infer steps within a complex motion or task and convert this into an FSM or STRIPS representation \cite{mataric,dillmann} using multiple training sessions. More recent systems allow refinements of a single demonstration through use of a simulator and editor \cite{cakmak}. Others attempt to build neural-net action models from passive observation of video \cite{Aloimonos-DNN}. While building such controllers is clearly useful, binding them to verbal commands is often an afterthought, and combination of taught elements is typically accomplished through a graphical interface \cite{code3} instead of via natural language. One notable exception is \cite{verbal-prog} where actions are targeted to certain objects called out through verbal referring expressions.

Several large-scale cognitive architectures include both learning and a natural language component. The ambitious LAAS-CNRS architecture \cite{sisbot-arch,sisbot-NL,SHARY} has dialog control, an ontology interface, spatial relation understanding, and anaphora resolution. Yet the emphasis here was on using language to set up planning problems, rather than as a conduit for acquiring new procedural knowledge. DIARC \cite{DIARC} is another complex architecture centered around a planner. It does an impressive job of extracting the parameterization of a new action, as well as pre- and post-conditions, directly from natural language. It can also be taught about command expansions, such as what ``follow me'' means \cite{DIARC-follow}. An enhanced version \cite{DIARC-learn} can learn sequences and object names, similar to \cite{Eli-AGI}. However it is largely confined to the deliberative action domain and it is unclear whether it could be extended to reactive routines or social conventions.

Our system is probably most similar in spirit to \cite{ASP}, except this uses temporal answer set programming instead of semantic nets and a production system. Overall, we present a relatively simple alternative to prior work, yet one that can still mediate several kinds of useful learning interactions, as will be described in the following sections.

\section{ALIA COGNITIVE ARCHITECTURE}
\label{sec:design}

%\noindent We are endeavoring to build a system where the top level behavior of an agent can be changed by simply talking to it. This enables one-shot learning by essentially just remembering (with appropriate generalization). This is in contrast to systems based on statistical machine learning, which may require hundreds of training examples to catch on. Quick learning is essential for one-off tasks where training effort cannot be amortized over a long service life. It is also useful for rapid customization (e.g., of robots) by end users. 

\noindent Our emphasis on acquiring procedural knowledge directly from natural language is what sets ALIA apart from many other cognitive architectures.
As a consequence of this, we have chosen a symbolic rule-based reasoner as the core of our system. 
Language most conveniently reduces to symbolic constructs, while reasoning and action recommendations are most often communicated in small chunks like rules. Fortunately, the brittleness of symbolic systems can be ameliorated by grounding the system in more robust statistical primitives, and 
the often chaotic control structure of rule-based systems can be tamed through a collection of mechanisms, as will be described later.

Fig.~\ref{fig:block} shows the overall structure of the implemented robot controller. Our reasoning subsystem is split into declarative and procedural parts, amplifying the dichotomy proposed in \cite{ACT-proc}. That is, there are inference rules that derive new facts, and separate policy operators that control action. Both of these can be taught directly by the user through the natural language interface. In operation, something akin to a goal is generally posted by the user to the attention buffer. A small amount of rule-based reasoning is done to elaborate the situation, then a policy operator is selected for execution in the action tree. To actually accomplish a task, the system relies on a grounding kernel that interfaces with the robot hardware.

\begin{figure}[t]
\centering
\includegraphics[width=0.7\columnwidth]{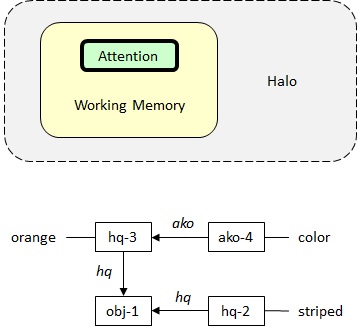}
\caption{Our system has three levels of memory (top). Attention is used to guide action, working memory establishes context, and the halo holds speculative deductions. Each level consists of semantic net assertions (bottom).} 
\label{fig:mem-levels}
\end{figure}

An important part of our solution is the three-level memory structure shown in Fig.~\ref{fig:mem-levels} (top). At the center is the attention buffer which holds recent facts and commands and is the primary conscious guide for the agent's activities. There can be multiple focal items active simultaneously. Once an item has been ``handled'' it is deactivated but persists in the attention buffer for a short amount of time. Outside attention there is working memory holding all the ancillary assertions linked to current or recently deactivated items. Finally, the outermost layer is the \textit{halo} which contains all those facts that are on the tip of your tongue but which only surface if needed. They remain unconscious in the sense that they cannot be used to make further deductions. 
%The interplay of these three layers in controlling autonomous behavior will be detailed in section \ref{sec:design} while a worked out example is given in section \ref{sec:tiger}.

The declarative representation used in our system is shown in Fig.~\ref{fig:mem-levels} (bottom), with nodes representing both objects and predicates. The actual names assigned to the nodes are irrelevant and are merely for debugging purposes. Predicate nodes have directed links marked by the role for each argument. Here \textit{hq} means ``has quality'' and \textit{ako} means ``a kind of''. Each node can have a belief value
associated with it as well as a lexical item (such as ``orange'' for \texttt{hq-3}) giving the system a strong Whorfian flavor. Predicates can also be modified by other predicates. For instance, the network depicted expresses not only that some object is orange, but also makes explicit that orange is a color. 

These networks are relatively simple, along the lines of \cite{Boris-NL} rather than more elaborate formalisms such as \cite{SNePS,AMR}. The overall system is largely concerned with understanding procedural sequences and interpreting imperative commands. There is no explicit model of time and hence no tense or modality annotations are required, unlike in narrative comprehension. Moreover, our networks are built on-demand and disposed of a short while later. They do not accrete into a large knowledge base over time that needs to be updated and checked for consistency. 

\begin{figure}[t]
\centering
\includegraphics[width=0.5\columnwidth]{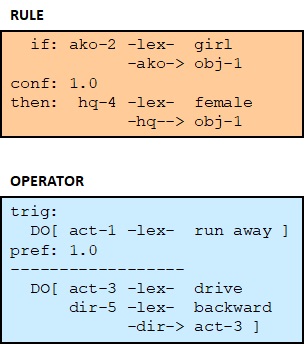}
\caption{Rules are deductions based on working memory with conclusions asserted in the halo. Operators are triggered by elements of the attention buffer and have a response composed of action directives.} 
\label{fig:rule-op}
\end{figure}

As shown in Fig.~\ref{fig:rule-op}, within the reasoning subsystem inference about facts is performed by rules (top), whereas advice about actions is conveyed by operators (bottom). The description of applicable situations is encoded as a semantic network, as are the details of the conclusions to be drawn or actions to be taken.
Each rule can have a degree of belief in its conclusion, while each operator can have a preference value associated with its selection. 
The examples here correspond to the natural language inputs ``Girls are (usually) female'' and ``To run away, (you could) drive backward'', where the parenthesized elements correspond to belief and preference, respectively. Suitable structured outputs are automatically derived from either text or speech -- they do not need to be entered in the intermediate form depicted here.

The reasoning engine works via forward chaining, where stimulus leads to response. This allows the system to spontaneously react to circumstances and exhibit initiative.
However unbridled chaining can lead to various sorts of runaway, such as counting. Suppose there is a rule that says ``if x then successor(x) = x + 1'' and we present the system with the number 2. We certainly do not want it to lock up while enumerating all the values up to infinity. There are related problems when computing deductive closure in systems derived from formal logic.
First, a lot of effort can be expended recomputing the world state every time some fact changes. 
Second, human-supplied rules are often only locally consistent (bounded rationality) and can lead to global contradictions despite still being usable. 

For this reason, we limit deduction to two steps.
Rules can only be matched against items in working memory, but then assert their conclusions in the halo.  
A version of concatenation \cite{MACROP} or chunking \cite{SOAR} can be used to get deeper lookahead, although this is often not necessary. Moreover, because the halo is conceptually ephemeral and is constantly being re-derived as the contents of working memory change, having a limited horizon saves substantial work.

Operators differ from rules in that they have intentional tags attached to particular nodes in the network. There are a number of such \textit{directives} including: NOTE, DO, ANTE, POST, CHK, FIND, ACH, KEEP, PUNT, and FCN. These differ primarily in how success is measured and how backtracking is controlled. For instance, a NOTE directive typically corresponds to a newly asserted fact. When this shows up in the attention buffer, the system selects an operator with a NOTE-based triggering condition to handle it. If the operator runs to completion, the top-level NOTE succeeds and is dismissed from the attention buffer. Control then passes to some other directive. If the operator fails, a different operator is tried next. If no operators remain, the top-level NOTE is still considered to have succeeded. By contrast, a DO directive usually corresponds to a command. As with NOTE, operators are tried successively until one succeeds. However, if there are no applicable operators, or all the selected ones fail, the DO directive itself fails.

Yet the DO operator is actually more complicated than NOTE because it gets automatically converted to the sequence ANTE-DO-POST. This allows the attachment of \textit{before} and \textit{after} methods. That is, when the system schedules a DO for execution it first attempts to run \emph{all} operators with matching ANTE triggers. Some of these may succeed while others fail, or there may be none to begin with. In any of these cases, only once all the possibilities are exhausted does control pass to the main DO directive. In this way the system makes sure that it has all the relevant information that might be needed before selecting an operator for DO. Similarly, no matter whether DO eventually succeeds or fails, once it is done all the applicable POST operators are called. This allows the robot to update its world state, or more gracefully recover from failures.

Unfortunately, production systems also suffer from interference between rules when the set grows large. One way around this is to partition the rules into a number of groups that are suited for different purposes. Following this approach, each of our operators has a trigger condition akin to a goal, and an enablement specification more like a typical antecedent. If the trigger does not match a suitable directive in the attention buffer, the operator is never considered for use.

Another important issue to consider is conflict resolution, when more than one operator is applicable to a situation. We first find antecedent matches where the facts are all above a certain belief threshold. This threshold can change over time, for instance if the robot is desperate to find some action to perform it can lower the threshold. Next we look at each operator's intrinsic preference to define a probability of selection. Here the raw values can be modulated by a function like exponentiation to shift the system's operating point along the explore-exploit continuum. For each operator above the minimum threshold, we combine this transformed value with the operator's specificity (as measured by the number of matching nodes) to get a final selection probability.

Arguably backwards chaining is also useful, particularly for responding to user-imposed goals. Yet language features, such as \texttt{cut} in Prolog, are often introduced to keep logic-based systems from needlessly chasing dead ends. Some systems separately authorize chaining direction on a rule-by-rule basis \cite{logic-FB}. Our approach is instead based loosely on \textit{determinations}~\cite{determine}. These are not exact implications but rather functional dependencies like ``if you know a person's nationality, you likely know the language he speaks''.
This piece of knowledge can be translated into the forward chaining advice ``if you want to know what language a person speaks, determine his nationality.'' A separate forward chaining set of rules would then capture the dependencies, such as ``if a person is from Brazil, he likely speaks Portuguese''. This is also reminiscent of the buffer filling then harvesting pattern found in EPIC \cite{EPIC}. The application of ALIA to perceptual reasoning using this approach is demonstrated in \cite{video-p}.

Finally, backward-chaining symbolic planning systems can end up agonizing over each step in deep plans. Yet, for the most part, humans prefer to use shallow reasoning to string together pre-built templates \cite{Schank}. That is, when going to the airport I do not have to think about opening the car door, sitting in the seat, turning on the ignition, etc. Instead, I simple ``drive the car''. For this reason, the consequents of operators are not restricted to atomic actions but can be whole sequential procedures with parallel paths and loops (cf. skills \cite{skills}). Specifically, the procedure part of an operator consists of a sequential \textit{chain} of \textit{plays}. A play can be as simple as a single directive, and it often is in the examples to be described in the next section. However, it can also be a set of parallel plays which must all complete before moving on to the next step in the chain. In addition, a play can have another non-required set of plays which are enabled while the required set is running. These auxiliary activities are automatically terminated, whether they are still running or not, when the main play completes.

The functioning and interplay of various directive types should become clearer during the discussion of the video demonstration, next.

\section{VIDEO DEMONSTRATION}

The video \cite{video} shows several examples of the user teaching the robot by talking to it. In this implementation the robot is fairly simple and the bulk of the processing (speech, reasoning, control) happens on a nearby Bluetooth-connected laptop. 
For sensing, the robot has a forward-facing triangulation-based optical rangefinder. This is a beam-like device mounted at the top of the gripper looking parallel to the ground which can sense most objects out to about 40cm. The robot also has a wifi-connected color camera, but it is not used here.
In terms of action, there are four degrees of freedom: two wheels, a gripper, and a lift stage. The associated grounding kernel responds to the commands drive, turn, grab/release, and raise/lower. Each of these is configured as a discrete instead of continuing action, although both modes could be present at the same time. That is, ``turn'' causes the robot to reorient by a fixed angle rather than spinning forever. The microcontroller on the robot takes care of generating pulses for the servos based on a serial communications line, but simple timeout-based action amounts are governed by the laptop. 

Although a typing interface also exists, the video demonstration has the user talking to the nearby laptop. The implementation runs under Windows 10 and uses Microsoft Speech Recognizer 8.0 through SAPI calls. Although this is a relatively old engine, it has the advantage of running locally so no network connection is required. Also, speech recognition still has some accuracy problems and can benefit from whatever additional constraints can be imposed. This particular engine allows speaker-dependent acoustic models and can be run in a mixed grammar/dictation mode. The use of grammar can not only limit the vocabulary to reasonable terms, but can also enforce preferred phrasing patterns. Thus, when the system is told to ``pick up the block'' by a speaker with a foreign accent, it \emph{never} erroneously hears ``peacock the block''. Although ``peacock'' is a valid English word, it obviously makes no sense in the robot deployment context. Such mistakes can be difficult to detect and repair after the fact.

\subsection{Procedures}

The most important advice a user can provide is how to accomplish some task, such as ``tidy up". The small robot shown does not actually perform this function but, as shown in the video,
it can be taught to dance through a sequence of verbal inputs:

\begin{center}
\textit{
To cha-cha drive forward then drive backwards \\
To shimmy turn left then turn right \\
To dance cha-cha then shimmy
}
\end{center}
 
\noindent When the user then says \textit{Please dance} the robot performs all four steps in sequence, something it was not able to do previously.

\begin{figure}[t]
\centering
\includegraphics[width=0.95\columnwidth]{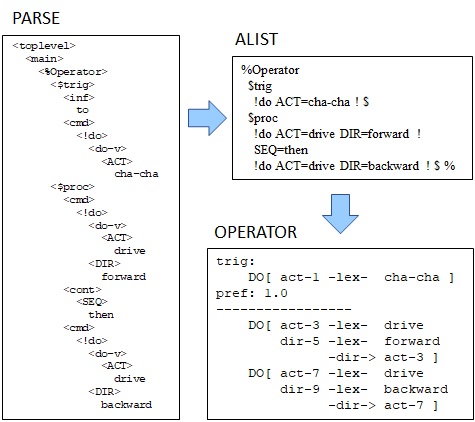} % Reduce the figure size so that it is slightly narrower than the column. Don't use precise values for figure width.This setup will avoid overfull boxes. 
\caption{The speech utterance \textit{To cha-cha drive forward then drive backwards} undergoes several steps of natural language processing to become a new operator.}
\label{parse}
\end{figure}

To process this sort of advice, the robot first listens to the user's speech as constrained by a Context Free Grammar. Figure~\ref{parse} shows the parse tree (left) resulting from the ``cha-cha" utterance. This is then digested into an association list (top right) consisting of a number of slots and values, along with constituent bracketing, by walking the tree and retaining selected nodes. Finally, the a-list gets converted to the internal representation (bottom right) of an operator which is added to the policy. 

The new operator is set up to match a DO directive where the triggering situation is an action associated with the lexical item ``cha-cha''. As described earlier, all operators that match the command directive are candidates for execution. However, this is modulated by exactness of match (calculated as the number of nodes in correspondence) and the operator's intrinsic suitability. Although this defaults to 1, it might be some other value depending on phraseology. For example, ``To cha-cha \textit{you could} ...'' would yield a preference of 0.8 while ``To cha-cha \textit{you must always} ...'' results in a preference of 1.2. These are user-supplied initial values but are subject to autonomous alteration by the system over the course of time. For instance, reinforcement learning could adjust them based on the success/failure of the suggested action, or they could be directly influenced by user utterances such as ``Hey, stop that!''.

The action part of the operator consists of a set of actions, here two DO directives. As can be seen, the actual description of what to do in each step is encoded as a semantic network. The first step is an action (\texttt{act-3}) associated with the word ``drive'' which is modified by a direction node (\texttt{dir-5}) associated with the word ``forward''. The second step is similar, except the direction is changed. The default is for the directives to be run sequentially, but the user can also specify concurrency by changing the conjunction used, e.g. ``and'' as opposed to ``then''. This does not make much sense for two drive commands, but it is useful in other circumstances. For instance, ``... drive forward and turn right'' causes the robot to move in an arc. Parallel execution is also particularly useful for saying something while performing an action. 

\begin{scriptsize}
\begin{verbatim}
      trig:   
          DO[ act-1 -lex-  drive 
              dir-1 -dir-> act-1 ]
      pref: 1.0
      ---------------
         FCN[ fcn-1 -lex-  base_drive 
                    -arg-> act-1 ]
\end{verbatim}
\end{scriptsize}

Each step is posted as a subgoal in the attention buffer and, in this case, will directly match a grounding function. This is a bridge between the symbolic system and the robot controller. For instance, the operator above matches any drive command having a direction but does not constrain what that should be. When run, this invokes a particular external function, (\texttt{base\_drive}) which has free access to the semantic network in the calling sequence. This function looks for words like ``backward'' as well as variants like ``backwards'' and can also examine other adverbial modification not explicitly listed in the trigger for the operator, such as ``slowly''. Using these parameters, it sets a speed for the robot and a timeout interval after which success is declared (and the velocity reverts to the default of zero). 

Note that while the grounding kernel automatically handles some variants like turning ``left'' or ``counterclockwise'' there may be things that were omitted, such as ``widdershins''. This can be handled either by specifying a new operator using the sentence ``To turn widdershins turn counterclockwise'' or, alternatively, using a macro-like halo rule ``Widdershins means counterclockwise''. Similarly, the grounding functions may have been written for the verb ``rotate'' instead of ``turn''. Here the aliasing rule approach is preferable. The user could tell the robot ``To turn left rotate left'' but would also have to define another operator for ``turn right'' (and possibly more if ``fast'' and ``slow'' were used). Instead, by simply saying ``turn means rotate'' a new lexical tag can be automatically added to the action node without changing any of the modifiers.

\begin{scriptsize}
\begin{verbatim}
        if: act-1 -lex- turn
      then: act-1 -lex- rotate
\end{verbatim}
\end{scriptsize}

In the video, the top-level user command ``Please dance'' posts a DO directive to the attention buffer. This is then matched to available operators and converted to a sequence of two more DO directives (cha-cha and shimmy). Each of these is resolved in turn to a sequence of more primitive DO directives, which are eventually matched to grounding operators like the one shown above. Note that grounding functions can also fail. For instance, if the robot knew its absolute position in the room and saw this was not changing, it could signal failure. At this point the architecture would backtrack and try a different operator for ``drive forward''. If none were found, it would backtrack still further to ``cha-cha'' and then to ``dance''.

\subsection{Reactions}

Advice is not limited to simple subroutine expansion but can also encompass reactive behaviors. In the home robot scenario, Freddy (the teenaged owner) might say ``Close the door whenever you see Mom coming". The video shows a similar situation in which the user tells the robot:

\begin{center}
\textit{
If something is very close then drive backwards
}
\end{center}

\noindent This behavior is not dependent on any top-level goal or command, but rather is operational all the time. What happens in the implemented system is that the robot continuously monitors its forward-facing rangefinder and tests whether the reading is less than 5cm. If so, the grounding kernel directly asserts the fact \textit{There is something very close} by posting the directive shown below to the attention buffer. This fact then automatically triggers suitable operators in the same manner as a command issued by the user. 

\begin{scriptsize}
\begin{verbatim}
      NOTE[  hq-1 -lex-  close
                  -hq--> obj-1
            deg-3 -lex-  very
                  -deg-> hq-1 ]
\end{verbatim}
\end{scriptsize}

If the robot is already working on some other command, this becomes an additional parallel action item. Generally, attentional foci that are posted more recently have a higher priority for their actions. The priorities are resolved within the grounding kernel. Suppose the robot was previously using its wheels to turn left, then this new backwards command takes precedence. However, if the ongoing action was merely to say something, there is no resource conflict and both actions would happen simultaneously.

\subsection{Prohibitions}

Just as important as what to do is what \textbf{not} to do. For instance, in the room cleanup scenario Freddy might want to say ``Never touch my sneakers!". In the video, the small robot is given a portion of Asimov's First Law:

\begin{center}
\textit{
You should never grab a person \\
but instead say I'm not allowed to
}
\end{center}

\noindent What this does is preempt the grabbing action while simultaneously generating a speech complaint, as encoded by the operator below.

\begin{scriptsize}
\begin{verbatim}
      trig:
        ANTE[ act-2 -lex-  grab
                    -obj-> obj-1
              ako-4 -lex-  person
                    -ako-> obj-1 ]
      ---------------
          DO[ act-6 -lex-  say
                    -obj-> txt-8
              txt-8 -str-  I'm not allowed to ]
        PUNT[ ]
\end{verbatim}
\end{scriptsize}

This illustrates the use of two more directives: ANTE and PUNT. As described earlier, a DO directive automatically cycles through three stages. It first gets converted to an ANTE directive and \textit{all} applicable operators are run. It then becomes the core DO directive and operators are tried until one succeeds or there are no more left, at which point it fails. Finally, it is converted to a POST directive and \textit{all} those applicable operators are run to properly clean up. The key to this prohibition is that it latches onto the factors considered before the main action is attempted. The other important part about this operator is the use of the PUNT directive. This causes the current chain to always fail and immediately initiates backtracking. Thus, when grabbing is being considered, this operator is triggered to first generate an utterance and then, when finished, to cause the top-level user-imposed DO directive to terminate.

Unfortunately, the current system cannot reliably assess whether an object is a human or not. However, it can be given some inference rules along with a fact about the object being grabbed, as demonstrated in the interaction:

\begin{center}
\textit{
If something is a girl it is a person \\
Mary is a girl \\
Grab Mary
}
\end{center}

The first sentence here results in a new halo rule:

\begin{scriptsize}
\begin{verbatim}
        if: ako-2 -lex-  girl
                  -ako-> obj-1
      then: ako-4 -lex-  person
                  -ako-> obj-1
\end{verbatim}
\end{scriptsize}

\noindent Using this, the declarative inference system automatically deduces the fact that Mary is a person, even though this is not directly evident from the top level ``grab" command. That is, the user cannot get away with an illegal command just because he failed to explicitly mention that Mary was a person.

\subsection{Permissions}

Often prohibitions are not general in nature, but tied to specific people, like permissions. For instance, Freddy might not like his pesky little sister snooping in his room and so tells the robot ``Do not show Janey where my phone is but instead tell the squirt to get lost". Again, the video shows an analog to this scenario with the small robot. Here, Rick is deemed to have lower authorization than other users. 

Suppose these two directives have been instilled:

\begin{center}
\textit{
If Rick tells you to do something don't but instead complain \\
To complain say I don't take orders from you
}
\end{center}

\noindent The first of these results in an ANTE operator to handle the ``don't'', similar to the grabbing example.

\begin{scriptsize}
\begin{verbatim}
      trig:
        ANTE[ act-7
              act-1 -lex-   tell
                    -agt--> obj-3
                    -dest-> obj-5
                    -cmd--> act-7 
              obj-3 -lex-   Rick
              obj-5 -lex-   you ]
      -----------------
          DO[ act-8 -lex-  complain ]
        PUNT[  ]
\end{verbatim}
\end{scriptsize}

The reason this operator will work as intended hinges on the practice of pre-pending a speech act to every user utterance. A command like ``Turn right'' actually posts a chain of two directives to the attention buffer as shown below.

\begin{scriptsize}
\begin{verbatim}
      NOTE[ input-4 -lex-   tell
                    -agt--> user-3  
                    -dest-> self-1  
                    -cmd--> act-5 ]
        DO[   act-5 -lex-   turn
              dir-7 -lex-   right
                    -dir--> act-5 ]
\end{verbatim}
\end{scriptsize}

Here the system first matches the NOTE directive to some relevant handler, something like the acknowledgement ``Okey Dokey'', before moving on to the DO directive. However, in this case, the core of the NOTE directive remains in working memory, recording the provenance of the command. Coupled with the known properties:

\begin{scriptsize}
\begin{verbatim}
      user-3 -lex- Rick
      self-1 -lex- you
\end{verbatim}
\end{scriptsize}

\noindent this is sufficient to trigger the newly supplied ``anti-Rick'' operator. As shown, this is composed of a DO directive to complain in some way followed by a PUNT directive, which functions as before. 

In this example we have also told the robot one way to complain: by saying ``I don't take orders from you''. However, there could be additional operators with different expansions, such as saying ``Forget it, puny human'' or even making a rude noise. When handling the DO operator, the system would choose randomly between these responses, assuming they all had the same preference. Notice that the main ``anti-Rick'' operator would not need to be re-taught -- all the variation is modularly contained within the generic directive to complain. Moreover, this complaining action could be reused in other circumstances without having to specify exactly how it should be accomplished each time.

The end result is that when a random user asks the robot to move, it complies. But when it knows that Rick is speaking, the robot digs in its heels.
Of course, preemptively refusing human orders could become problematic in the long term.

\section{DISCUSSION}

While ALIA can do simple factual reasoning and knows how to string procedures together, it should not be used to do everything. Instead, the intent is for the grounding functions to do most of the heavy lifting, with natural language used for run-time scripting. In particular, the grounding functions can be arbitrarily complex, such as a DNN-based object recognizer (invoked by ``find the cereal box") or a vSLAM-based navigation system (invoked by ``go to the kitchen''). Furthermore, although ALIA itself is largely a symbolic system, there is no reason that these libraries have to be. The combination, in fact, has some interesting synergies where the ``slow'' ALIA system enables migration of some tasks to the ``fast'' grounding system \cite{Kahneman}.

For instance, the robot might be told ``The bottle to the left of the green box is Advil". Using linguistic interpretation and primitive object segmentation, it can determine what portion of the image corresponds to mentioned item. But now, since it also has the label ``Advil'', it can submit the pair as a training example to a DNN. This could in turn help the robot more directly spot the item amidst clutter in the future. This same approach of verbal guidance of attention then labeling can be used in other areas, as well. For instance, the user could say ``To get to Tim's office, go out the door, turn left, then enter the third door on your right". Again, ALIA would follow these symbolic directions to traverse one particular path to the destination. However, meanwhile vSLAM could be running in the background building a map of the environment the robot was passing through, and bind the designation ``Tim's office" to the endpoint. Over time this should eventually let the robot get to Tim's office from a variety of different starting locations using conventional path planning.

In the big picture, humans rarely discover new things or invent new processes. Rather, the majority of what they do and what they know is passed through culture, and the prime vector for this is language. Enabling learning from language allows robots to share in this culture. Users can teach the robot what to think, do, and say. The implemented examples here show how a user can interact with the robot to:

\begin{itemize}
\item give explicit commands (e.g. ``drive forward")
\item guide it in performing some task (e.g. ``to cha-cha")
\item instruct it how to respond to events (e.g. ``very close")
\item explain how the world works (e.g. ``a girl is a person")
\item lay down ethical constraints (e.g. ``don't grab")
\item communicate notions of trust (e.g. ``Rick")
\item impart a personality (e.g. ``complain")
\end{itemize}

\noindent This demonstrates how powerful language can be, and how it can help bootstrap AI agents closer to human-like behavior.

\bibliography{biblio}
\end{document}